\documentclass[12pt]{article}
\usepackage{amsmath,amssymb}
\usepackage{graphicx,psfrag,epsf}
\usepackage{float, subfig}
\usepackage{enumerate}
\usepackage{multirow}
\usepackage{natbib}
\usepackage{url} 
\usepackage{xcolor}

\newcommand{\blind}{0}

\date{}

\begin{document}

\def\spacingset#1{\renewcommand{\baselinestretch}%
{#1}\small\normalsize} \spacingset{1}


\if0\blind
{
  \title{\bf Aggregated Pairwise Classification of Statistical Shapes}
  \author{Min Ho Cho, Sebastian Kurtek, and Steven N. MacEachern\\
    Department of Statistics, The Ohio State University}
  \maketitle
} \fi

\if1\blind
{
  \title{\bf Aggregated Pairwise Classification of Statistical Shapes}
  \maketitle
} \fi

\bigskip
\begin{abstract}
The classification of shapes is of great interest in diverse areas ranging from medical imaging to computer vision and beyond. While many statistical frameworks have been developed for the classification problem, most are strongly tied to early formulations of the problem - with an object to be classified described as a vector in a relatively low-dimensional Euclidean space. Statistical shape data have two main properties that suggest a need for a novel approach: (i) shapes are inherently infinite dimensional with strong dependence among the positions of nearby points, and (ii) shape space is not Euclidean, but is fundamentally curved. To accommodate these features of the data, we work with the square-root velocity function of the curves to provide a useful formal description of the shape, pass to tangent spaces of the manifold of shapes at different projection points which effectively separate shapes for pairwise classification in the training data, and use principal components within these tangent spaces to reduce dimensionality. We illustrate the impact of the projection point and choice of subspace on the misclassification rate with a novel method of combining pairwise classifiers.
\end{abstract}

\noindent%
\vfill

\newpage
\spacingset{1.45} 
\section{Introduction}
\label{sec:intro}

Classification of shapes is a fundamental task in many application areas where the primary data object is an image. For example, in medical imaging, radiologists and doctors are often interested in classifying patients to different disease types based on shapes of anatomical structures. Consequently, the statistical analysis of shape data, and in particular the shape classification task, are of great interest to the research community. Our focus in this paper is on the multiclass shape classification problem, which presents some unique challenges. To elucidate the main difficulties, we begin by explaining what we mean by ``shape data."

The literature on shape analysis has considered many different mathematical representations of shape, including finite point sets or landmarks \citep{dryden1998statistical,dryden1992size,cootes}, level sets \citep{Malladi}, skeletal models \citep{Pizer2013}, and diffeomorphic transforms or deformable templates \citep{grenander-miller:98,Glaunes2008}, among others.  Stretching the definition of a landmark, consider a set of 2D images of leaves, each marked with a dense set of landmarks.  The landmarks provide the outline of a leaf, and with them its shape.  If the image is shifted, rescaled, or rotated, the shape remains unchanged; other transformations change the shape. \cite {kendall1984shape} recognized these invariances and defined shape as the geometric information in the set of landmarks that remains when translation, scaling and rotation have been filtered out.  

In many applications, such as the leaf example that we return to in Section \ref{sec:empirical}, it seems most natural to study the shape of an object via its entire outline rather than through a finite set of landmarks.  In the 2D setting which we focus on, the shape is a planar, closed curve.  The functional representation of such a curve replaces the set of landmarks with an alternative description.  The curve is parameterized by a starting point on the shape and a mapping from the unit interval that describes the traversal of the shape, ending the journey at the starting point.  Early versions of the functional representation relied on an arc-length parameterization for the traversal \citep{zahn-roskies:72,klassen2004analysis}. However, later papers showed that the arc-length parameterization was too rigid \citep{srivastava2011shape,kurtek2012statistical,srivastava2016functional}, and that statistical analysis of shapes benefits from the more flexible elastic deformations. These elastic parameterizations rely on registration to determine the optimal point-to-point correspondences across objects (we describe this process formally later).

In this work, we adopt the popular square-root velocity function (SRVF) representation for elastic shape analysis of planar, closed curves \citep{joshi2007novel,srivastava2011shape}. There are two main advantages associated with this framework: (1) the SRVF simplifies a specific instance of an elastic metric \citep{mio2007shape,younes1998computable} to the simple $\mathbb{L}^2$ metric, facilitating efficient computation, and (2) the SRVF shape space is a quotient space of the unit Hilbert sphere for which many geometric quantities of interest have analytic expressions. These two ingredients allow parameterization-invariant (in addition to the other standard shape preserving transformations) comparisons and statistical models of shape. We exploit this representation for the model-based shape classification task. We provide more mathematical details on the SRVF and the formal definition of the associated shape space in Section \ref{sec:SRVF}.

\subsection{Motivation}

The curved nature of the shape space prevents us from directly using standard techniques for classification that are strongly tied to Euclidean geometry. The normal distribution, for example, is defined in $\mathbb{R}^d$, with extension to spaces of infinite dimension provided by the Gaussian process. Linear Discriminant Analysis (LDA) and Quadratic Discriminant Analysis (QDA), two popular model-based classification techniques which we consider in this work, are described in various fashions, but are intrinsically tied to the normal distribution, and hence to Euclidean spaces. While alternative classification approaches exist, usually based on shape distances and nearest neighbor-type classification rules \citep{kurtek2012statistical,6411702}, we argue that a model-based approach provides more flexibility in the definition of multiclass classification procedures. In particular, the ability to compute likelihoods for different shape classes allows us to aggregate multiple pairwise classifiers in a principled manner.

One standard approach to classification of shapes is based on ``linearization'' of the shape space \citep{pal2017riemannian,srivastava2011shape}. This is done by choosing a particular point in the shape space, usually given by the overall sample mean, identifying the (linear) tangent space at this point, and projecting the shapes into the tangent space via the inverse-exponential map.  The role of the inverse-exponential map is illustrated in Figure~\ref{fig:tang}. Once the shapes are projected into the tangent space, one can apply standard classification techniques. A major drawback of such an approach is that a single tangent space is generally chosen for the pairwise and multiclass problems. In the multiclass case, if one or more populations are very far from the others, projecting all shapes into a single tangent space at the overall mean introduces significant distortion into the tangent space shape coordinates. In turn, this has a negative effect on any subsequent statistical task, e.g., classification.

A second major challenge in shape classification is the high dimensionality of the shape space. Theoretically, the shape space and corresponding tangent space are infinite-dimensional since we use a functional representation of shape.  In practice, the outlines are represented using a fine discretization, typically in the order of hundreds to thousands of points for an individual shape. This discretization leads to a tangent space of large, but finite dimension. The large dimension necessitates modification of LDA and QDA through dimension reduction or regularization. We pursue dimension reduction via a standard form of tangent Principal Component Analysis (tPCA), and modify the LDA and QDA classification procedures accordingly.

\begin{figure}[!t]
\begin{center}
    \includegraphics[scale=.35]{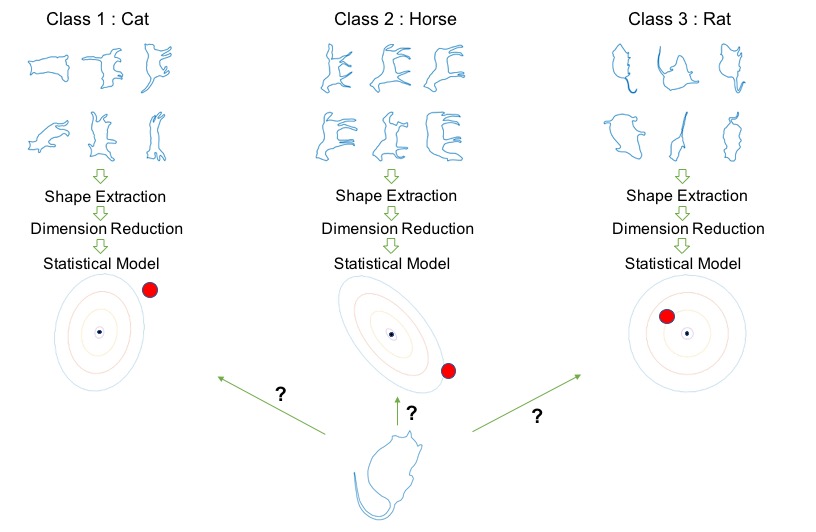}
\end{center}
    \vspace{-.2in}
    \caption{A general outline of the multiclass shape classification problem.}
    \label{fig:classify}
\end{figure}

Figure \ref{fig:classify} summarizes the multiclass shape classification problem. We are given labeled training data in the form of curves. The first task is to extract the shape from these curves. This requires within class registration as mentioned earlier. The second task is to linearize the shape space to a Euclidean space, reduce the dimension, and then fit a classwise statistical model. The models we consider are based on LDA and QDA. Finally, to classify a new unlabeled object, one has to register its outline to each class and make an assignment based on a classification rule; we consider likelihood-based classification.

\subsection{Contributions}

Our investigations show the need to modify the standard linearization approach to classification of shapes. The main reason for this is the arbitrariness in the choice of the tangent space where the full procedure is carried out. While the sample mean leads to a small distortion in inter-point distances after projection into the tangent space (see Section 7.4.1 in \cite{srivastava2016functional}), this does not guarantee improved classification performance for the multiclass problem. In other words, since the lower-dimensional Euclidean coordinates of the shapes (after tPCA dimension reduction) are intimately tied to the tangent space used to define them, the chosen point of projection can have a major impact on classification performance. This is especially true when the training data has large dispersion, making a single tangent space approximation unsuitable for statistical analysis. In view of this, we list our main contributions.
\begin{itemize}


\item For multiclass shape classification, we provide a heuristic that suggests different projection points for different pairwise problems. We develop a method to combine the pairwise results, and compare its performance to a classifier based on a single projection point. The new procedure has substantially better performance than the currently-used single projection methods.
\item For aggregating the pairwise problems, we first propose a one shot method that chooses the class with the highest likelihood (based on the LDA or QDA models). Additionally, we define a recursive approach that drops the class with the lowest likelihood, and then recomputes all relevant quantities without this class present in the data. This procedure is especially effective when there are several classes that differ greatly from most groups in the data.
\item Finally, we suggest an intermediate method that is also based on aggregation of pairwise problems but only uses a single tangent space. This alternative procedure performs better than the one shot method in a single tangent space. As expected, it does not perform as well as the new pairwise procedure described above. The main motivation for this intermediate approach comes from the large computational cost of working with all pairwise tangent spaces (this requires the computation of all pairwise means), when the number of classes is large.

\end{itemize}

The rest of this paper is organized as follows. Section \ref{sec:prelim} briefly reviews the SRVF framework for elastic shape analysis of planar curves and defines tools for computing all relevant statistics. Section \ref{sec:classify} begins by describing the currently used one shot classification approach in a single tangent space. It then defines the three novel procedures, which rely on pairwise statistics and dimension reduction to different degrees. Section \ref{sec:simulation} provides a simple simulation study that motivates the use of pairwise procedures in classification. Section \ref{sec:empirical} includes detailed empirical studies of a popular plant leaf dataset as well as an animal dataset with very diverse shapes. Section \ref{sec:discuss} provides a short discussion and lays out some directions for future work.

 
\section{Elastic Shape Analysis Preliminaries}
\label{sec:prelim}

We review the elastic shape analysis framework, based on the square-root velocity function representation, proposed by \cite{srivastava2011shape}. For brevity, we do not provide all details of this framework, and refer the interested reader to \cite{srivastava2016functional}.

\subsection{Shape Preserving Transformations}

Let $\beta : D \rightarrow \mathbb{R}^2$ represent a parameterized, planar curve. For an open curve $D=[0,1]$, while for a closed curve $D=\mathbb{S}^1$. We restrict our analysis to the set of absolutely continuous curves. While our focus is on classification of planar shapes, these methods are also applicable to higher-dimensional curves. Under our notation, $\beta(t) = (x(t), y(t))$ where $x,y$ are scalar-valued coordinate functions, and $t$ is a parameter that traces the path $\beta(t)$, in $\mathbb{R}^2$, from the start point of the curve to the end point of the curve. In the case of a closed curve, the start and end points are exactly the same. 

\begin{figure}[!t]
\begin{center}
\begin{tabular}{cc}
    \begin{minipage}{.45\textwidth}
      \includegraphics[width=.9\linewidth]{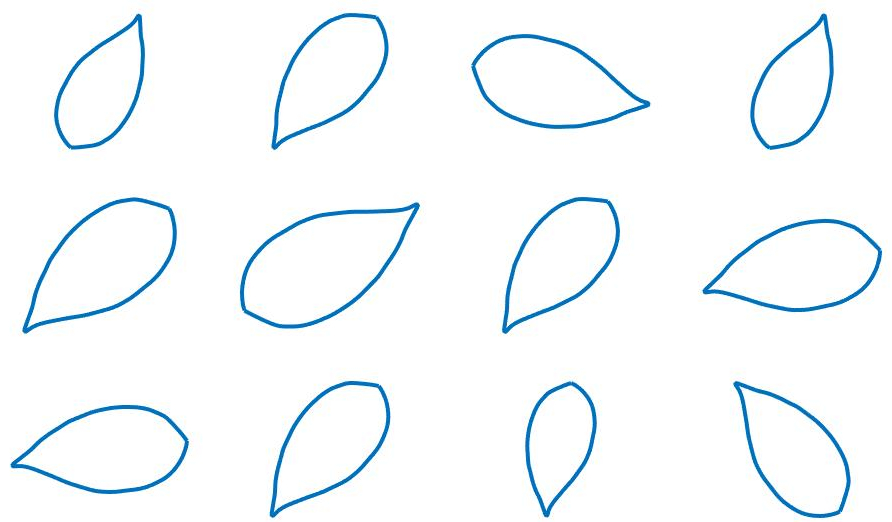}
    \end{minipage}
    &
    \begin{minipage}{.45\textwidth}
      \includegraphics[width=\linewidth]{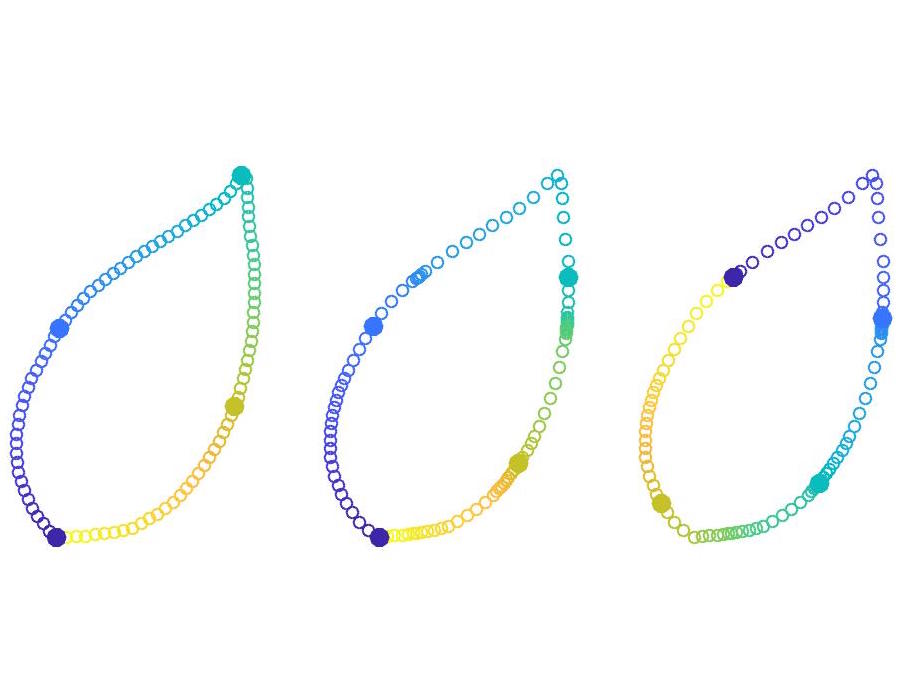}
    \end{minipage}
   \\
\end{tabular}
\end{center}
\vspace{-0.5in}
\caption{Illustration of shape preserving transformations. The same leaf shape with different translations, rotations and scales (left), and different parameterizations (right).} \label{fig:invariance}
\end{figure}

As mentioned in the introduction, shape is invariant to translation, scaling, rotation and re-parameterization. The translation of a curve $\beta$ is given by $\beta+T$ (this addition is carried out for each parameter value $t$), where $T\in\mathbb{R}^2$. Invariance to this transformation is easiest to achieve, as it will be imposed directly in the representation. The rescaling of a curve $\beta$ is given by $c\beta$, where $c\in\mathbb{R}_{+}$. Invariance to scale can be achieved via normalization: we will consider unit length curves only. The rotation of a curve $\beta$ is given by $O\beta$ (again, this matrix multiplication is applied for each parameter value $t$); here, $O\in SO(2)$ where $SO(2)=\{O\in\mathbb{R}^{2\times 2}|O^TO=OO^T=I,\det(O)=1\}$ is called the special orthogonal group. Finally, the re-parameterization of a curve $\beta$ is given by composition $\beta\circ\gamma$, where $\gamma\in\Gamma$ and $\Gamma$ is the set of orientation preserving diffeomorphisms of $D$. In constrast to translation and scaling, rotation and re-parameterization are not easy to filter out of the representation. 

Figure~\ref{fig:invariance} illustrates all of the shape preserving transformations. The left panel shows the effect of translation, rotation and rescaling. The right panel shows the effect of re-parameterization. In particular, we display the sampling of points on the curve imposed by the different parameterizations using colors: the start point is given in black and the parameterization traces the curves from purple to yellow. The filled-in points correspond to the same parameter values across the three curves.

\subsection{Square-Root Velocity Function Shape Space and Distance}
\label{sec:SRVF}

Comparison of shapes of different objects is fundamental to shape analysis. This task, as well as subsequent statistical tasks, requires the definition of a distance between shapes. The $\mathbb{L}^2$ norm given by $\|\beta_1 - \beta_2\|=\sqrt{\int_D |\beta_1 (t)-\beta_2 (t)|^2 dt}$, where $|\cdot|$ denotes the Euclidean norm in $\mathbb{R}^2$, seems a natural choice. However, this distance is not parameterization invariant because $\| \beta_1 - \beta_2 \|$ is not guaranteed to equal $\| \beta_1 \circ \gamma - \beta_2 \circ \gamma \|$ \citep{srivastava2011shape}. This suggests the need for a different distance on shapes.

\cite{mio2007shape} defined a family of elastic Riemannian metrics that is invariant to all of the aforementioned shape preserving transformations, including re-parameterization. These metrics are called elastic because they provide a natural interpretation of shape deformations in terms of their bending and stretching/compression. However, despite these nice mathematical properties, their practical use in shape analysis was limited due to computational difficulties until \cite{joshi2007novel} and \cite{srivastava2011shape} introduced the square-root velocity function (SRVF)

\begin{equation}
q(t) \equiv \left\{
\begin{array}{cc}	
     \dot{\beta}(t) /\sqrt{|\dot{\beta}(t)|} 	& \text{if } |\dot{\beta}(t)| \ne 0\\
     0 			& \text{otherwise}
\end{array}\right.
\end{equation}

\noindent where $\dot{\beta}$ is the derivative of $\beta$ with respect to $t$. The SRVF simplifies the elastic metric to the $\mathbb{L}^2$ metric, thereby facilitating easy computation \citep{srivastava2011shape}. If $\beta$ is absolutely continuous, then its SRVF $q$ is square-integrable, i.e., an element of $\mathbb{L}^2(D,\mathbb{R}^2)$, henceforth referred to simply as $\mathbb{L}^2$ \citep{robinson2012functional}. Further, one can uniquely recover the curve $\beta$ from its SRVF $q$, up to a translation, via the relation $\beta(t)=\int_0^t q(s)|q(s)|ds$, where $t=0$ is the start point of the parameterization. For the remainder of this paper, we focus on shape analysis of curves facilitated by the SRVF transform.

The translation of a curve is automatically filtered out under the SRVF representation, as it is based on the derivative of the curve. Restricting the set of all absolutely continuous curves to those that have unit length results in a unit norm constraint on the associated SRVFs: $\int_D |\dot{\beta}(t)| dt = \int_D |q(t)|^2 dt = 1$. Thus, we define the pre-shape space as $\mathbb{S}_\infty = \{ q\ |\ \|q\|^2 = 1 \}$, the unit sphere in $\mathbb{L}^2$. We call this the pre-shape space as we have not yet filtered out rotations and re-parameterizations. Under the $\mathbb{L}^2$ metric, the distance between $q_1,q_2\in \mathbb{S}_\infty$ is given by $d(q_1,q_2)=\cos^{-1}(\langle q_1,q_2\rangle)$, where $\langle q_1,q_2\rangle=\int_D q_1(t)^T q_2(t) dt$.

Next, we aim to unify the representation of all SRVFs that are within a rotation and/or re-parameterization of each other. To do this, we first note that the SRVF of a rotated curve, $O\beta$, is simply given by $Oq$. The SRVF of a re-parameterized curve, $\beta\circ\gamma$, is given by $(q,\gamma)=(q\circ\gamma)\sqrt{\dot{\gamma}}$, where again $\dot{\gamma}$ denotes the derivative. Briefly returning to the topic of invariance, it is precisely the extra term $\sqrt{\dot{\gamma}}$ that makes the $\mathbb{L}^2$ metric under the SRVF representation invariant to re-parameterizations. We define equivalence classes of the type $[q]=\{O(q, \gamma) ~|~ O \in SO(2), \gamma \in \Gamma \}$ and deem all SRVFs within an equivalence class to have the same shape. In other words, each equivalence class represents a unique shape. The resulting shape space, given by all such equivalence classes, is $\mathcal{S} = \mathbb{S}_\infty/(SO(2) \times \Gamma) = \{ [q] ~|~ q \in \mathbb{S}_{\infty} \}$.

The distance between two shapes is defined as the distance between their equivalence classes. To define this distance, we use the $\mathbb{L}^2$ metric defined on the pre-shape space:

\vspace{-0.15in}
\begin{equation}\label{geoddist}
d_{\mathcal{S}}([q_1],[q_2])= \min_{O \in SO(2), \gamma \in \Gamma} \cos^{-1}( \langle q_1, O(q_2, \gamma) \rangle).
\end{equation}

\begin{figure}[!t]
\begin{center}
\begin{tabular}{|c|c|}
\hline
    \begin{minipage}{.45\textwidth}
      \includegraphics[width=\linewidth, height=20mm]{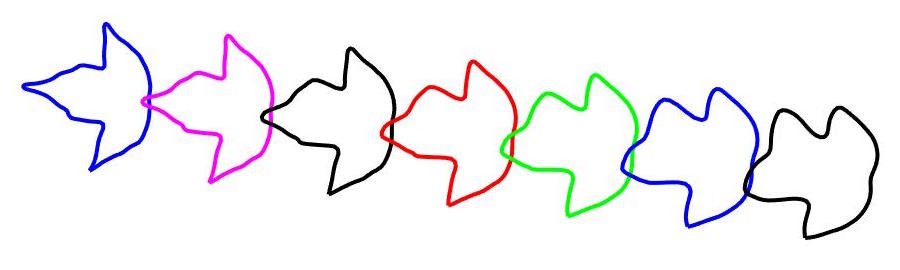}
    \end{minipage}
&    \begin{minipage}{.45\textwidth}
      \includegraphics[width=\linewidth, height=20mm]{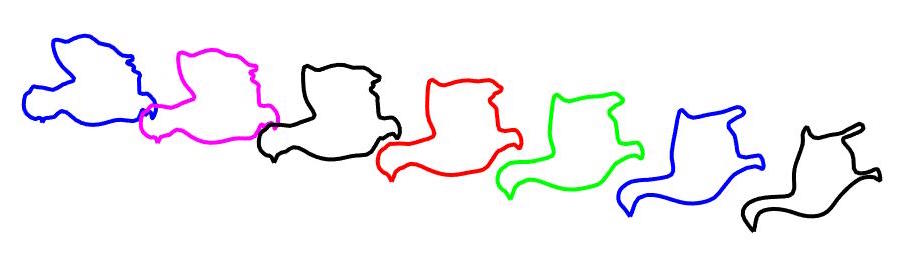}
    \end{minipage}
   \\
$d_\mathcal{S}=0.4794$ & $d_\mathcal{S}=0.6684$ \\ \hline
\end{tabular}
\end{center}
\vspace{-.25in}
\caption{Elastic geodesic paths, and corresponding distances, between different shapes.} \label{fig:geodesic}
\end{figure}

\noindent The optimization problem over $SO(2)$, also called Procrustes rotation, is solved using singular value decomposition (Section 5.7 in \cite{srivastava2016functional}). To solve for the optimal re-parameterization one can either use a gradient descent approach (Section 5.8 in \cite{srivastava2016functional}) or the Dynamic Programming algorithm \citep{robinson2012functional}. In the case of closed curves, one must additionally perform an exhaustive search for the optimal starting point on the shape. The optimal rotation and re-parameterization (minimizers of Equation (\ref{geoddist})) solve the registration problem. After registration, one can construct a geodesic path between two shapes by connecting them via a great circle on the pre-shape space $\mathbb{S}_{\infty}$. Figure \ref{fig:geodesic} shows two examples of elastic shape geodesics between different types of objects. The length of each path is precisely the shape distance $d_\mathcal{S}$. Visually, the elastic geodesics represent natural deformations between shapes.

\subsection{Projection onto Tangent Space and Dimension Reduction}

\begin{figure}[!t]
    \centering
    \includegraphics[scale=.4]{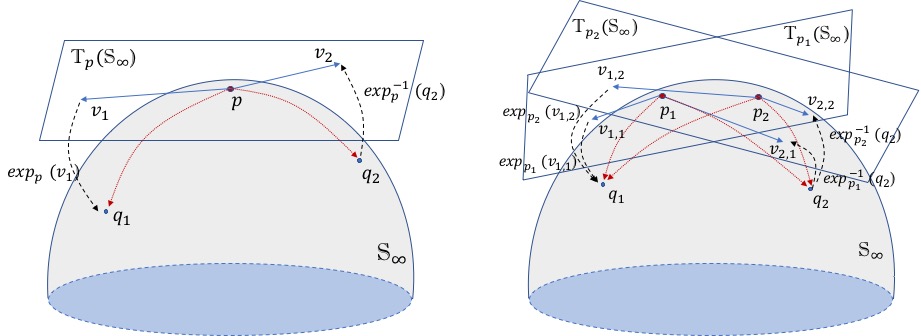}
    \caption{Left: Two SRVFs $q_1, q_2$ and their corresponding tangent space representations $v_1,v_2$, computed using the inverse-exponential map at a point of projection $p$. Right: Two SRVFs $q_1, q_2$ and different tangent space coordinates $v_{1,1},v_{2,1}$ and $v_{1,2},v_{2,2}$, respectively, obtained using the inverse-exponential map at two different points of projection $p_1$ and $p_2$.}
    \label{fig:tang}
\end{figure}

To facilitate classification methods naturally designed for linear spaces, such as LDA or QDA, we introduce the exponential map and its inverse to linearize the elastic shape space. Since the pre-shape space is a unit sphere, the expressions for these mappings are analytic. At a given point $p \in S_{\infty}$, there is a (linear) tangent space, $T_p(S_{\infty})$. The exponential map $\exp_p : T_p(\mathbb{S}_{\infty}) \rightarrow \mathbb{S}_{\infty}$ is given by (for $p\in\mathbb{S}_{\infty}$ and $v\in T_p(\mathbb{S}_{\infty})$)
\begin{equation}
\exp_p(v) = \cos(\|v\|)p + \sin(\|v\|)(v/\|v\|),
\end{equation} 
\noindent where $\|\cdot\|$ is the $\mathbb{L}^2$ norm. This expression can be used to map vectors (along geodesics) from a tangent space to the representation space. The inverse of this mapping, called the inverse-exponential map $\exp^{-1}_{p} : \mathbb{S}_{\infty} \rightarrow T_{p}(\mathbb{S}_{\infty})$, is given by (for $p,q\in\mathbb{S}_{\infty}$)
\begin{equation}\label{invexp}
\exp^{-1}_{p}(q) = v = (\theta / \sin (\theta))(q - \cos (\theta) p),
\end{equation}
\noindent where $\theta = \cos^{-1}(\langle q_1, q_2 \rangle)$; this expression can be used to transfer points from the representation space to a linear tangent space. We use the inverse-exponential map to transfer the shapes from the nonlinear shape space to an approximating (linear) tangent space. The left panel of Figure \ref{fig:tang} provides an illustration of the exponential and inverse-exponential maps. The right panel shows that the tangent space coordinates depend heavily on the point used to define the tangent space. This, in turn, can have a major impact on statistical tasks performed in the tangent space, including classification. 

Most commonly, the tangent space chosen for statistical shape analysis is defined at the sample mean shape \citep{le2001locating}, called the sample Karcher mean, which is defined using the shape distance $d_{\mathcal{S}}$ given in Equation (\ref{geoddist}) (for data $q_1,\dots,q_n\in\mathbb{S}_{\infty}$): 
\begin{equation}
[\bar{q}] = \arg\min_{[q] \in \mathcal{S}} \sum^n_{i=1} d_{\mathcal{S}}([q],[q_i])^2.
\end{equation}
\noindent While this mean is an entire equivalence class, we simply select one element of it for subsequent analysis, i.e., we choose some $\bar{q}\in[\bar{q}]$. The specific $\bar{q}$ that is chosen has no impact on the subsequent analysis.

The computation of the Karcher mean involves an optimization problem which is solved using a gradient descent approach (see, e.g., \cite{kurtekcviu} or \cite{dryden1998statistical}). Given a mean shape, we can define the Karcher covariance and study variability within and across shape classes using tangent Principal Component Analysis (tPCA). As a first step, we project all of the data into the tangent space at the mean using $v_i = \exp^{-1}_{[\bar{q}]}(q^*_i) \in T_{[\bar{q}]}(\mathcal{S}),\ i=1,\ldots,n$, where $\exp^{-1}$ is given in Equation (\ref{invexp}), $q_i^*=O^*(q_i,\gamma^*)$, and $O^*$ and $\gamma^*$ are the rotation and re-parameterization of $q_i$, respectively, that minimize $d_{\mathcal{S}}([\bar{q}],[q_i])$. In principle, there is a sample covariance for the vectors $v_i$ in the infinite dimensional tangent space. In practice, the curves are sampled using a finite number of points, say $m$. Thus, one can simply form the observed tangent data matrix $V \in \mathbb{R}^{n \times 2m}$ (by stacking the $x,y$ coordinates for each $v_i$ into a long vector of size $2m$), and then compute the Karcher covariance matrix, $Q \in \mathbb{R}^{2m \times 2m}$, using $Q=(1/(n-1))V^TV$. Note that for typical shape data $n<<2m$. The LDA and QDA classification approaches rely on covariance matrices, which are singular in this setting. This necessitates dimension reduction. 

\begin{figure}[!t]
\begin{center}
\begin{tabular}{|c|c|c|}
\hline
(a)&(b)&(c)\\
\hline
    \begin{minipage}{.2\textwidth}
      \includegraphics[width=\linewidth]{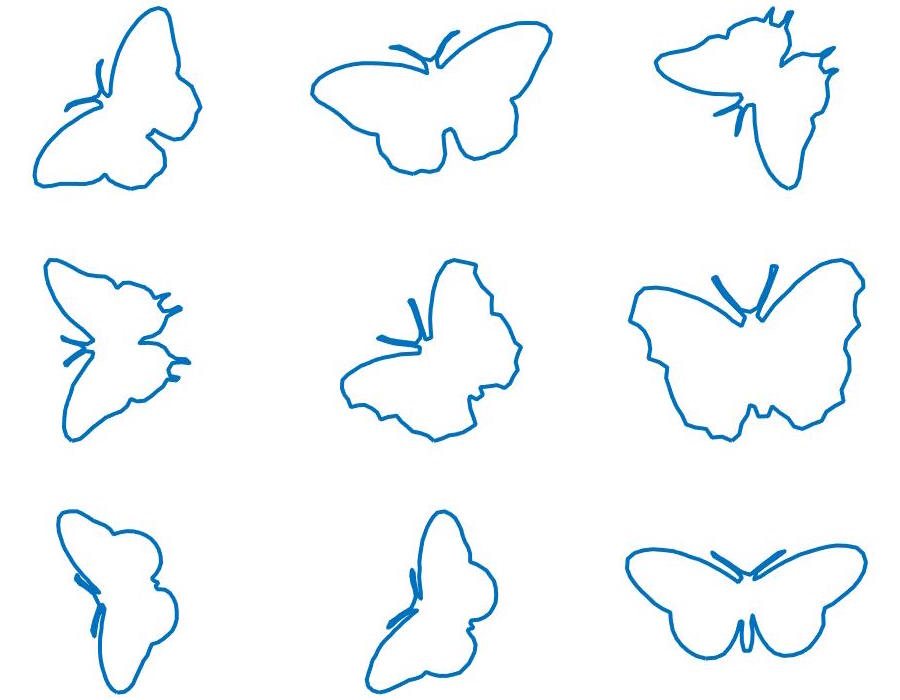}
    \end{minipage}
    &
    \begin{minipage}{.5\textwidth}
      \includegraphics[width=\linewidth]{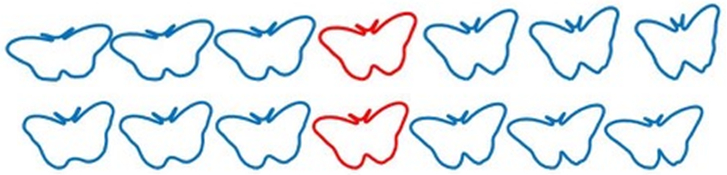}
    \end{minipage}
    &
    \begin{minipage}{.2\textwidth}
      \includegraphics[width=\linewidth]{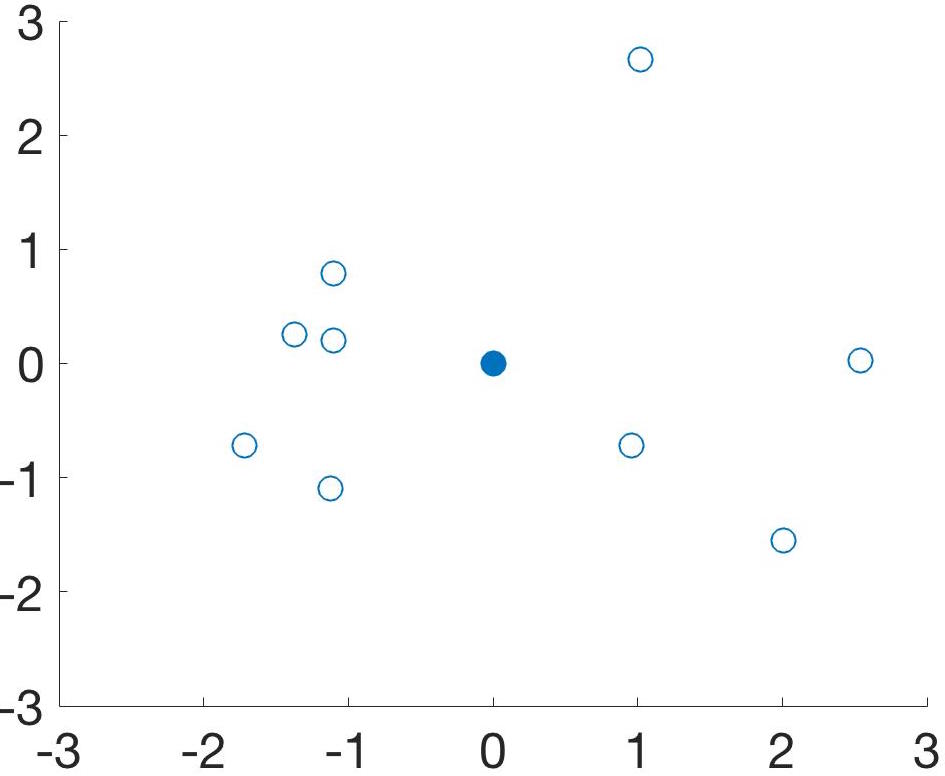}
    \end{minipage}
   \\
   \hline
\end{tabular}
\end{center}
\vspace{-.2in}
\caption{(a) A sample of nine butterfly outlines. (b) First two elastic principal directions of shape variability in the sample (mean shape in red). (c) Scatterplot of the 2D PC coefficients computed in the tangent space at the mean shape (filled-in point).} \label{fig:mean}
\end{figure}

While one can potentially apply any common statistical technique for dimension reduction to the data matrix $V$, we use the tPCA approach \citep{dryden1998statistical,kurtek2012statistical}. First, we use singular value decomposition to compute $Q=U\Sigma U^T$, where $U$ is an orthonormal matrix with columns specifying the principal directions of shape variation, and $\Sigma$ is a diagonal matrix with non-negative entries arranged in decreasing order specifying the principal component variances. Selecting $r < n-1$, one has a lower-dimensional Euclidean representation of the shapes in the tangent space as $C\in\mathbb{R}^{n\times r}$, with $c_{ij}=v_iU_j,\ i=1,\dots,n,\ j=1,\dots,r$. These PC data are used for tangent space classification of shapes with LDA or QDA. Figure \ref{fig:mean} shows an example of averaging and dimension reduction for a sample of nine butterfly shapes. The computed mean shape and first two principal directions of variability are natural summaries of the sample shapes. We also show a projection of the mean (filled-in point) and the data (hollow points) onto these first two principal directions, i.e., a plot of the first two PC coefficients.


\section{Classification of Shapes on Tangent Spaces}
\label{sec:classify}

We describe four different procedures for classification of shapes using Linear Discriminant Analysis (LDA) and Quadratic Discriminant Analysis (QDA) in tangent spaces. The first approach is in current use and serves as a baseline. We also discuss practical considerations in the context of data analysis. Throughout this section, we assume that the training data is balanced across classes. The unbalanced case can be handled using reweighting when computing the sample mean shape and by weighting the aggregated likelihoods.

\subsection{One Shot Classification on a Single Tangent Space}
\label{sec:meth1}

In this baseline approach to classification \citep{pal2017riemannian,srivastava2011shape,kurtek2012statistical}, we begin by computing the overall mean shape $\bar{q}$, and a PC coefficient representation of the shapes in the tangent space at $\bar{q}$, using training data pooled over all $K$ classes. Since the linearized shape data often have large dimension compared to the amount of available training data, we use tPCA for dimension reduction arriving at $r$ dimensions. Under the assumption of normality in this $r$-dimensional space, the log-likelihood of a new observation $x$ under QDA is given by   
\begin{equation}
l_{\bar{q}}(x; \hat{{\bf \mu}}_k, \hat{\Sigma}_k) = -\frac{1}{2} \log |2 \pi \hat{\Sigma}_k | - \frac{1}{2} ( x-\hat{\mu}_k)^T  \hat{\Sigma}^{-1}_k (x-\hat{ \mu}_k),
\end{equation}
\noindent where $\hat{\mu}_k$ and $\hat{\Sigma}_k$ are the mean and covariance estimated using training data in the PC coefficient space of class $k$. For LDA, we use the pooled estimate of the $r \times r$ covariance matrix, $\hat{\Sigma}_P = \frac{1}{K} \sum^{K}_{k=1} \hat{\Sigma}_k$, in place of each $\hat{\Sigma}_k$. After computing $l_{\bar{q}}({\bf x}; \hat{{\bf \mu}}_k, \hat{\Sigma}_k)$ for each class, we choose the class with the largest log-likelihood. In this method, we use a single projection point and a single PC space for dimension reduction; for brevity, we refer to this approach as SS. Furthermore, we use a ``one-shot'' (OS) decision for classification. 



\subsection{Aggregated Pairwise Classification on Single Tangent Space}
\label{sec:meth2}

In the multiclass case, one can improve upon the SS-OS approach, especially when one class of shapes is very different from the others. The unusual class may be easy to identify, and yet plays a significant role in determining the lower-dimensional PC space. As a result, the estimated PCs may be ineffective in discriminating between the other classes, leading to a higher than needed misclassification rate. This motivates us to introduce an approach based on PC decompositions of all possible pairwise covariance matrices.

Under this approach, we find $\bar{q}$, the mean shape of the training data pooled over all $K$ classes. The shapes are then projected into the tangent space $T_{[\bar{q}]}(\mathcal{S})$ using the inverse-exponential map. For each pair of classes, $i<j,\ i=1,\dots,K,\ j=1,\dots,K$, PCs are extracted from the covariance matrix computed using training data in classes $i$ and $j$ only. All training data are then represented in terms of these PCs and the pairwise log-likelihood of a new observation $x$ (based on the LDA or QDA model), $l^{i,j}_{\bar{q}}(x; \hat{\mu}_k, \hat{\Sigma}_k)$, is computed. The $M= {K \choose 2}$ log-likelihoods are aggregated by taking the mean $\bar{l}_{\bar{q}}(x; \hat{\mu}_k, \hat{\Sigma}_k) = M^{-1} \sum_{i < j} l^{i,j}_{\bar{q}}(x; \hat{{\bf \mu}}_k, \hat{\Sigma}_k)$. Use of the geometric mean of the likelihoods (arithmetic mean of the log-likelihoods) has a long history in Bayesian statistics and has been used for, e.g., combining expert opinion \citep{berger2013statistical}, combining partial Bayes factors \citep{berger1996intrinsic}, and synthesizing different analyses \citep{yu2011bayesian}. The new observation is then assigned to the class with the maximum mean log-likelihood (OS approach). In this method, we use a single tangent space and pairwise PC spaces for dimension reduction; for brevity, we refer to this approach as SP. In later sections, we demonstrate that this approach can provide significant improvements in classification over the SS-OS method.

\subsection{Aggregated Classification on Pairwise Tangent Spaces}
\label{sec:meth3}

The projection from the nonlinear shape space to the linear tangent space distorts inter-shape distances. The amount of distortion depends on multiple factors including the point of projection and the dispersion of the data. Pursuing the heuristics of pairwise comparisons by projection to the tangent space at the sample Karcher mean, we consider projections to all pairwise tangent spaces, followed by aggregation.

Under this approach, for each pair of groups in the training data, $i<j$, we compute the pairwise mean shape, $\bar{q}_{i,j}$, and determine the tangent space, $T_{[\bar{q}_{i,j}]}(\mathcal{S})$. We estimate PCs in $T_{[\bar{q}_{i,j}]}(\mathcal{S})$ using the training data in classes $i$ and $j$ only. Then, data from all classes are projected into these pairwise tangent spaces, and PC-based means and variances are estimated. This leads to the log-likelihood of a new observation $x$ (based on the LDA or QDA model), $l_{\bar{q}_{i,j}}(x; \hat{{\bf \mu}}_k, \hat{\Sigma}_k)$, defined in each pairwise tangent space. The multiple log-likelihoods are then combined as before using $\bar{l}_{\bar{q}_{\cdot,\cdot}}(x; \hat{\mu}_k, \hat{\Sigma}_k) = M^{-1} \sum_{i < j} l_{\bar{q}_{i,j}}(x; \hat{{\bf \mu}}_k, \hat{\Sigma}_k)$. The new observation is then assigned to the class with the maximum average log-likelihood (OS approach). In this case, we use pairwise tangent spaces and pairwise PC spaces for dimension reduction; for brevity, we refer to this approach as PP. In many cases, this procedure further improves upon the SP-OS method. 


\subsection{Aggregated Pairwise Classification with Recursion}
\label{sec:meth4}

The distortion induced by a projection point far from a pair of classes can lead to a very small and numerically unstable contribution to the aggregated log-likelihood. If severe enough, the classification rule can be destabilized. The impact of these poor projection points (and PC spaces) can be limited through use of a recursive procedure. We begin with the calculation of $\bar{l}_{\bar{q}_{\cdot,\cdot}}(x; \hat{\mu}_k, \hat{\Sigma}_k)$ for each class $k$ and new observation $x$. For the recursion, the class with the smallest mean log-likelihood is identified and dropped, leading to a similar problem with $K-1$ classes. The recursion continues with a succession of problems with fewer classes until a single class remains.

As an example, suppose there are $K$ classes. In the first stage, there are $M_1 = {K \choose 2}$ different tangent spaces defined at projection points $\{ \bar{q}_{i,j}, i < j\}$. The mean log-likelihood for class $k$ is $\bar{l}_{\bar{q}_{\cdot,\cdot}}(x; \hat{\mu}_k, \hat{\Sigma}_k) = M^{-1}_1 \sum_{i < j} l_{\bar{q}_{i,j}}(x; \hat{{\bf \mu}}_k, \hat{\Sigma}_k)$. If $\bar{l}_{\bar{q}_{\cdot,\cdot}}(x; \hat{\mu}_{k_1}, \hat{\Sigma}_{k_1})$ (where $k_1 \in \{ 1,\ldots,K \}$) is the smallest among all mean log-likelihoods, class $k_1$ is dropped from the comparison. Then, we re-aggregate all of the log-likelihoods without projections associated with class $k_1$. In the second stage, there are $M_2 = {K-1 \choose 2}$ projection points that do not include class $k_1$: $\{ \bar{q}_{i,j}, i < j,\ i,j \ne k_1 \}$. The new mean log-likelihood for class $k$ and new observation $x$ is given by $\bar{l}_{\bar{q}_{\cdot,\cdot}}(x; \hat{\mu}_k, \hat{\Sigma}_k) = M^{-1}_2 \sum_{i < j,\ i,j \ne k_1} l_{\bar{q}_{i,j}}(x; \hat{{\bf \mu}}_k, \hat{\Sigma}_k)$. If $\bar{l}_{\bar{q}_{\cdot,\cdot}}(x; \hat{\mu}_{k_2}, \hat{\Sigma}_{k_2})$ is the smallest re-aggregated log-likelihood among those of all $k$ except $k_1$, class $k_2$ is dropped from the comparison. Again, we re-aggregate the log-likelihoods without projections involving classes $k_1$ and $k_2$, and repeat this procedure. After repeating it $K-1$ times, we obtain a unique class for the final classification decision. This recursive approach (REC) can be used instead of the OS method described earlier for SP and PP. In general, REC does not provide the same classification decision as OS. 



\subsection{Practical Considerations}

\begin{figure}[!t]
    \centering
    \includegraphics[scale=.35]{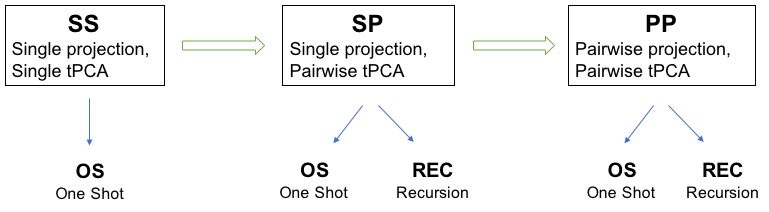}
    \vspace{-.1in}
    \caption{A diagram of different shape classification methods.}
    \label{fig:flow}
\end{figure}

In Sections \ref{sec:meth2} and \ref{sec:meth3}, we described two methods for multiclass shape classification that use pairwise procedures for dimension reduction to different degrees. Furthermore, in Section \ref{sec:meth4}, we proposed a recursive approach for aggregating pairwise classification results. Figure~\ref{fig:flow} provides a flowchart of all of the approaches, starting with the baseline method SS and progressing to increasingly pairwise procedures (SP and PP). For SP and PP, the user has an additional choice of using the OS or REC decision for classification.


 
In general, in terms of classification accuracy, PP outperforms SP, which outperforms the baseline SS. However, the three methods have different computational costs, which must be taken into consideration when choosing the best approach for a given multiclass shape classification problem. The main computational bottleneck is the search for the sample Karcher mean, which is performed using an iterative, gradient-based algorithm. The SS and SP approaches require only one such computation, albeit with a potentially large sample size, since they rely on a single tangent space approximation. In contrast to SS which uses a single PC space for classification, SP estimates all pairwise PC spaces, making it computationally more expensive. However, this increase in computational complexity is minimal. On the other hand, PP requires ${K \choose 2}$ computations of the sample Karcher mean, making it much more computationally expensive, especially when $K$ is large. Thus, there is a tradeoff between computational cost and classification accuracy. In addition, for SP and PP, the final classification decision based on aggregated likelihoods can be made via the OS or REC approaches. The OS method is simple and easy to interpret. But, if the training data contains a mix of very similar and very diverse classes, then the REC procedure helps classification performance by eliminating the worst classes in a stepwise manner. 

Finally, there are two additional choices in these classification procedures: (1) LDA vs.\ QDA, and (2) the dimensionality of the PC space. The first choice has been widely explored in the past for various problems, and we do not discuss it here further. The second choice is nontrivial, and there is no single prescription that applies across different problems and datasets. In general, we aim to achieve a low-dimensional yet faithful Euclidean representation of shape data via PCs.  We have found that the proposed approaches are very robust to these two choices. This is confirmed in Section \ref{sec:empirical} for real data examples. 
 

\section{Simulation Study}
\label{sec:simulation}

We construct a one-dimensional example that conveys the effect of nonlinear projections on multiclass classification performance. We define three classes, each having a $t$-distribution with a common scale parameter but different location parameters. First, we sample data from these distributions. Next, the sampled data are transformed using a CDF-inverse CDF pair that contracts the tails of the distribution. This step is analogous to the nonlinear projections used for classification of shapes. Then, we apply standard LDA for classification; for a new observation, the classification decision rule chooses the class with the closest sample mean of the transformed data. Specifically, let $y_{ik},\ i=1,\dots,n,\ k=1,2,3$, be the $i$th random observation drawn from a $t$-distribution with $\nu = 5$ degrees of freedom centered at $\mu_k$, i.e., $y_{ik}\sim t_5 (\mu_k)$. We transform $y_{ik}$ to $x_{ik}$ using $x_{ik} = \Phi^{-1} (F(y_{ik}-q))$, where $F$ is the CDF of the $t_5(0)$ distribution, $\Phi$ is the CDF of the standard normal distribution, and $q$ is a center point for the transformation.

\begin{figure}[!t]
\begin{center}
\begin{tabular}{|c|c|c|}
\hline
(a)&(b)&(c)\\
\hline
    \begin{minipage}{.3\textwidth}
      \includegraphics[width=\linewidth, height=40mm]{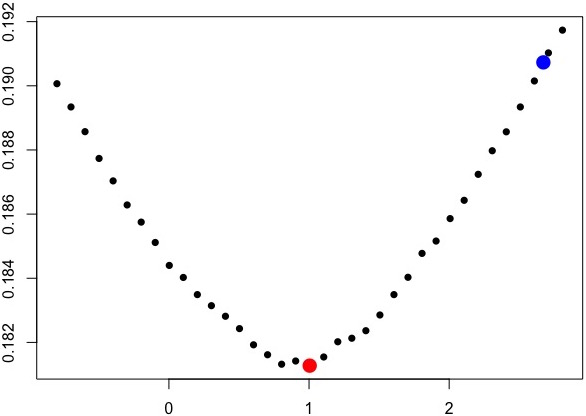}
    \end{minipage}
    &
    \begin{minipage}{.3\textwidth}
      \includegraphics[width=\linewidth, height=40mm]{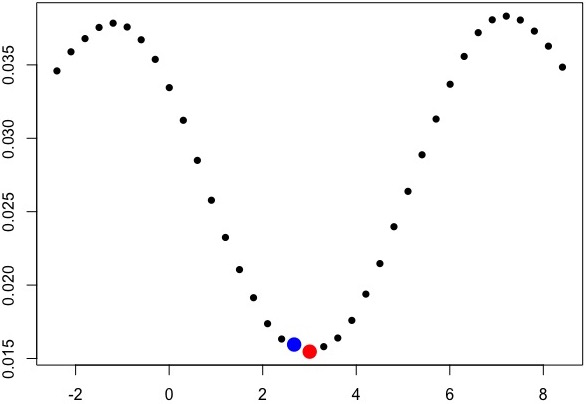}
    \end{minipage}
    &
    \begin{minipage}{.3\textwidth}
      \includegraphics[width=\linewidth, height=40mm]{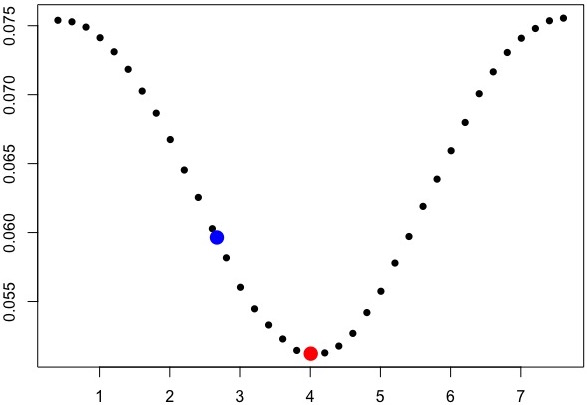}
    \end{minipage}
   \\
   \hline
\end{tabular}
\end{center}
\caption{Misclassification rates for different choices of $q$ with performance using the pairwise mean in red and at the overall mean in blue: (a) class 1 vs.\ class 2, (b) class 1 vs.\ class 3, (c) class 2 vs.\ class 3.}
\label{fig:qgrid}
\end{figure}

We begin by sampling random $y_{ik}$ with $n=100,000$ in each of the three classes and $\mu_1 = 0,\ \mu_2 = 2,\ \mu_3 = 6$. To investigate the effect of the point $q$ on classification performance, we generate the transformed data $x_{ik}$ using various choices for $q$, and compare classification performance based on the overall closest mean and pairwise closest means. Figure~\ref{fig:qgrid} displays the results of this simulation. The left, middle and right panels show results of pairwise classification for class 1 vs.\ class 2, class 1 vs.\ class 3 and class 2 vs.\ class 3, respectively. The values of $q$ are given on the $x$-axis, while the misclassification rates are on the $y$-axis. The black points mark the misclassification rate for different choices of $q$, the blue point marks the misclassification rate when the overall mean (across all three classes) is used as the center point for the transformation, and the red point marks the misclassification rate when the pairwise mean is used as the center point for the transformation. Use of the pairwise means for classification is superior to use of the overall mean. For example, in (a), we are trying to discriminate between class 1 with $\mu_1 = 0$ and class 2 with $\mu_2 = 2$. The pairwise mean for this case is equal to $1$ (red point) while the overall mean is equal to $2.67$ (blue point). As one moves away from the pairwise mean in either direction, the classification performance deteriorates severely. Aggregating results across the three pairwise problems results in an average misclassification rate of $0.09$; the overall mean approach yields a misclassification rate of $0.17$, which is nearly double. The first simulation shows the benefits of choosing pairwise projection points for classification.


Next, we assess four different LDA approaches: (i) LDA based on a single transformation at the overall mean, (ii) pairwise LDA with OS to determine class assignment, (iii) LDA at the optimally chosen transformation point with OS to determine class assignment, and (iv) pairwise LDA with REC to determine class assignment. For method (iii), we choose from 21 evenly spaced candidate points starting at the smallest mean and ending at the largest mean. Instead of selecting fixed shift parameters $\mu_k$, we randomly sample them from a uniform distribution on $[0,10]$, i.e., $\mu_{k}  \sim Unif(0,10),\ k=1,2,3$. For each classification experiment, we again use $n=100,000$.

\begin{figure}[!t]
\begin{center}
\begin{tabular}{|c|c|c|}
\hline
(a)&(b)&(c)\\
\hline
    \begin{minipage}{.3\textwidth}
      \includegraphics[width=\linewidth, height=50mm]{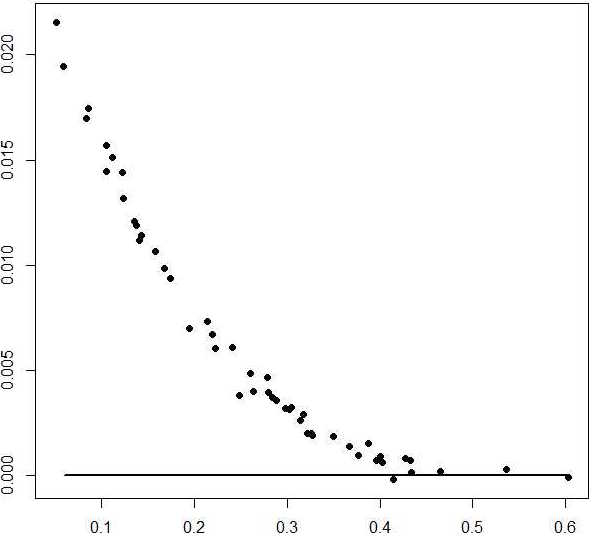}
    \end{minipage}
    &
    \begin{minipage}{.3\textwidth}
      \includegraphics[width=\linewidth, height=50mm]{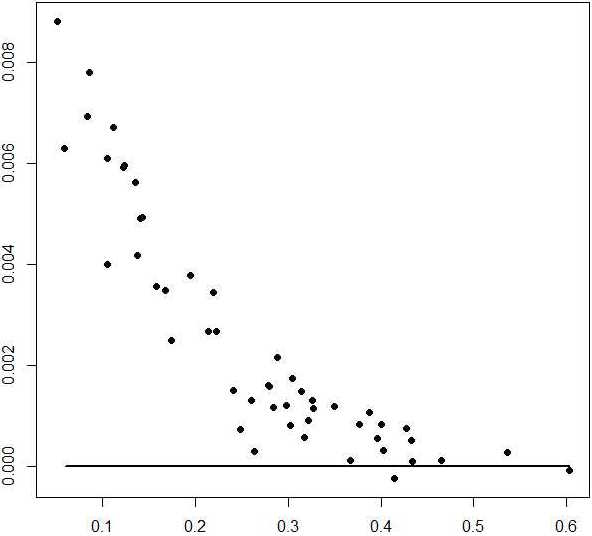}
    \end{minipage}
    &
    \begin{minipage}{.3\textwidth}
      \includegraphics[width=\linewidth, height=50mm]{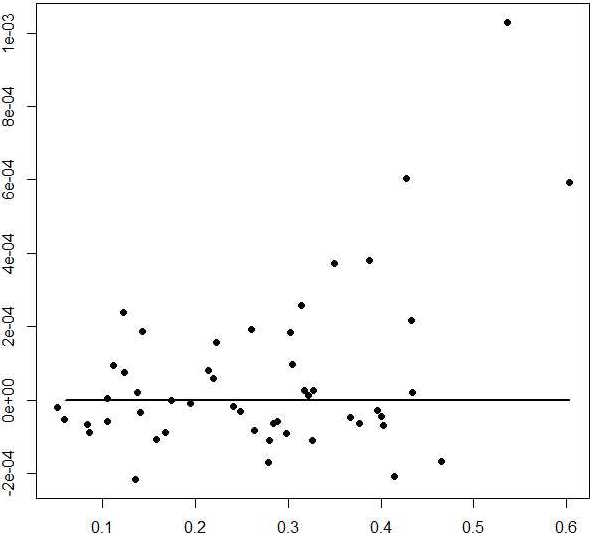}
    \end{minipage}
   \\
   \hline
\end{tabular}
\end{center}
\caption{Excess misclassification rates ($y$-axis) of (a) method (i), (b) method (ii) and (c) method (iii) as compared to the misclassification rate of method (iv) ($x$-axis).}
\label{fig:analog}
\end{figure}

Figure~\ref{fig:analog} shows 50 replicates (using the above-described settings) for three different comparisons: (a) method (i) vs.\ method (iv), (b) method (ii) vs.\ method (iv), and (c) method (iii) vs.\ method (iv).\ In each scatterplot, the $x$-axis gives the misclassification rate of method (iv) while the $y$-axis gives the excess misclassification rate defined as the misclassification rate of methods (i), (ii) or (iii) minus the misclassification rate of method (iv). The solid line represents no excess misclassifciation. Points located above the solid line favor method (iv). In panels (a) and (b), method (iv) is clearly superior for easier classification problems. When the misclassification rate reaches approximately $0.4$ or $0.5$, there appears to be no benefit from using REC as the decision rule as opposed to OS. Panel (c) demonstrates that the recursive method is as good as method (iii), which uses a search for the optimal transformation point $q$. The one-dimensional nature of this simple simulation allows for such a search. Since our focus is on classification of shapes, which lie in an infinite dimensional space, such a search becomes extremely computationally expensive, and is perhaps even impossible.

\section{Empirical Studies}
\label{sec:empirical}

We apply the proposed approaches to multiclass classification of two real shape datasets: plant leaves and animals. First, we consider a problem with a relatively small number of classes, by selecting only a few species from the leaf data. Then, we consider the entire datasets of leaves and animals. We begin with a brief description of the two datasets.

\subsection{Data Description}

We first work with the Flavia Plant Leaf dataset\footnote{\url{http://flavia.sourceforge.net/}} \citep{wu2007leaf}. The closed outlines used in our work were extracted from images of plants captured using a digital camera. The entire dataset consists of 1,907 observations of plant leaves split into 32 different classes corresponding to the species of the plants.

\begin{figure}[!t]
\begin{center}
\begin{tabular}{|c|c|}
\hline
(a)&(b)\\
\hline
    \begin{minipage}{.3\textwidth}
      \includegraphics[width=\linewidth]{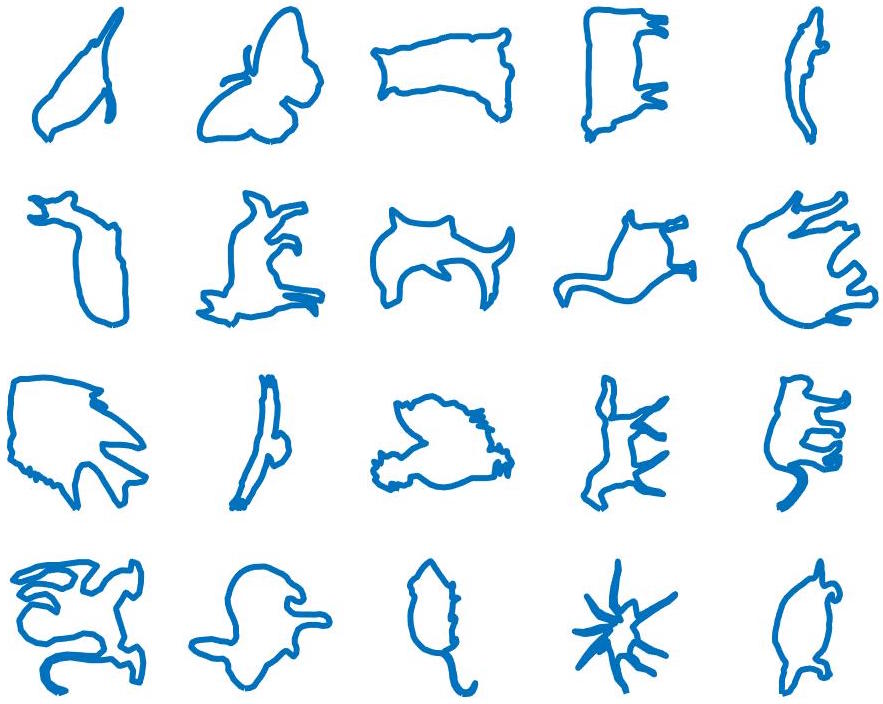}
    \end{minipage}
    &
    \begin{minipage}{.3\textwidth}
      \includegraphics[width=\linewidth]{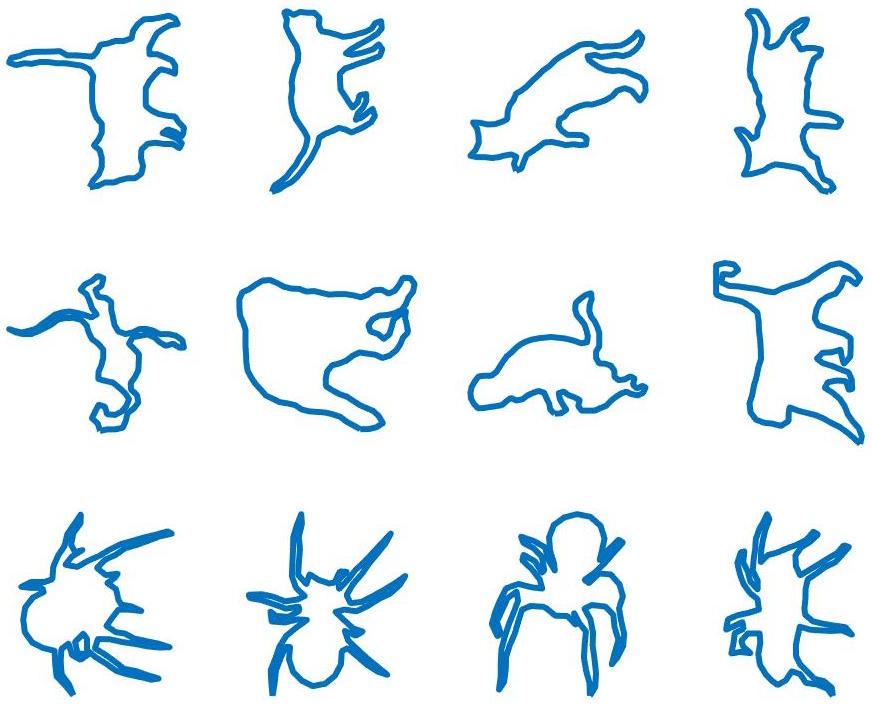}
    \end{minipage}
   \\
   \hline
\end{tabular}
\end{center}
\vspace{-.25in}
\caption{(a) One sample of each animal. (b) Examples of cats, monkeys and spiders.}
\label{fig:animals}
\end{figure}

\cite{bai2009integrating} provide shape data for different animals whose outlines were segmented from natural images. The entire dataset consists of 100 observations for 20 different types of animal: bird, butterfly, cat, cow, crocodile, deer, dog, dolphin, duck, elephant, fish, flying bird, chicken, horse, leopard, monkey, rabbit, mouse, spider, tortoise. In Figure~\ref{fig:animals}(a), we show a single example for each animal class (in the same order as the above list). In Figure ~\ref{fig:animals}(b), we show a few examples of cats, monkeys and spiders. We note that there is a lot of pose variability within each class, making the classification problem very difficult.

\subsection{Classification of a Small Subset of the Leaf Dataset}
\label{sec:fourleaves}

\begin{figure}[!t]
\begin{center}
\setlength{\tabcolsep}{0pt}
\begin{tabular}{cccc}
\includegraphics[scale=.187]{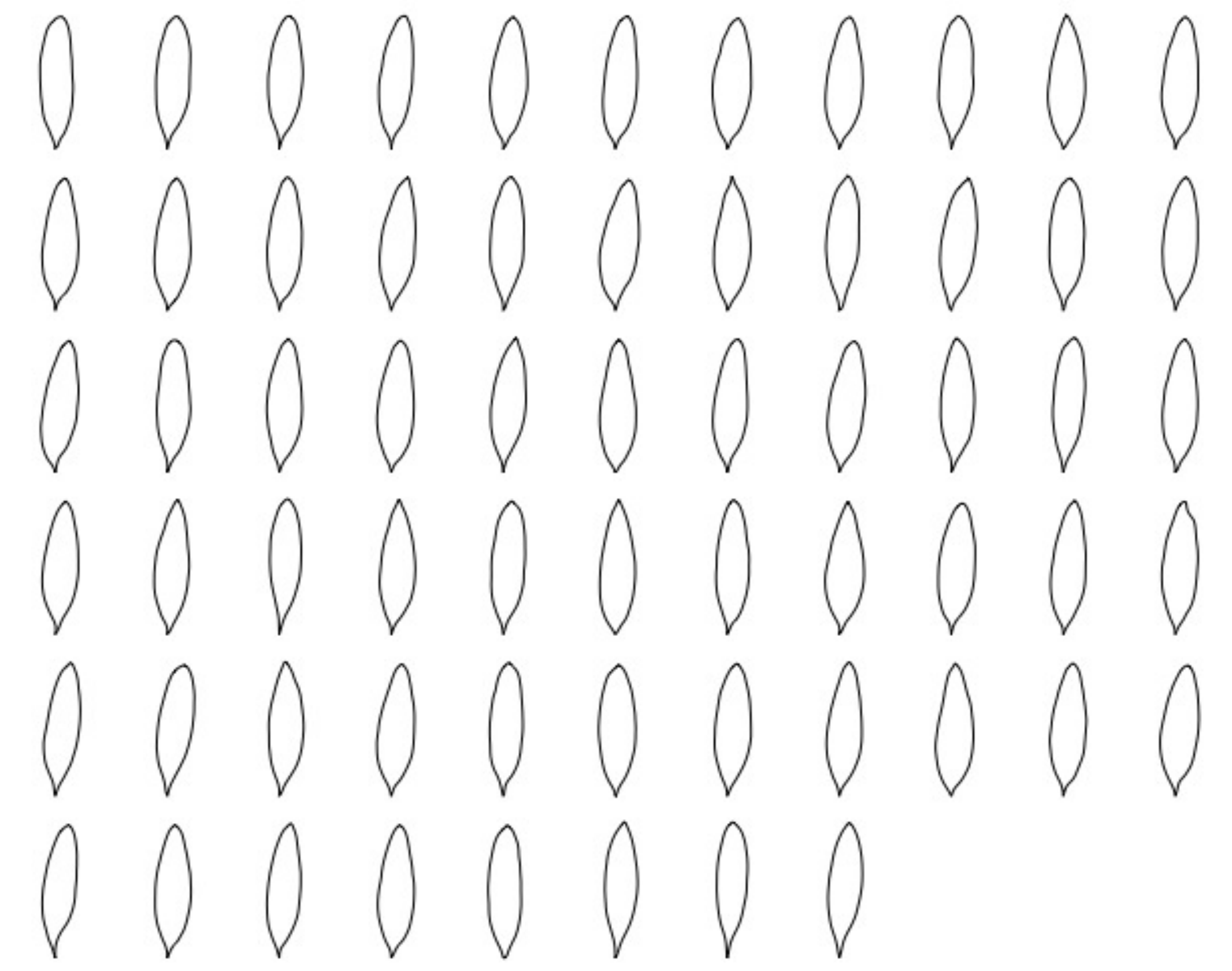}&\includegraphics[scale=.187]{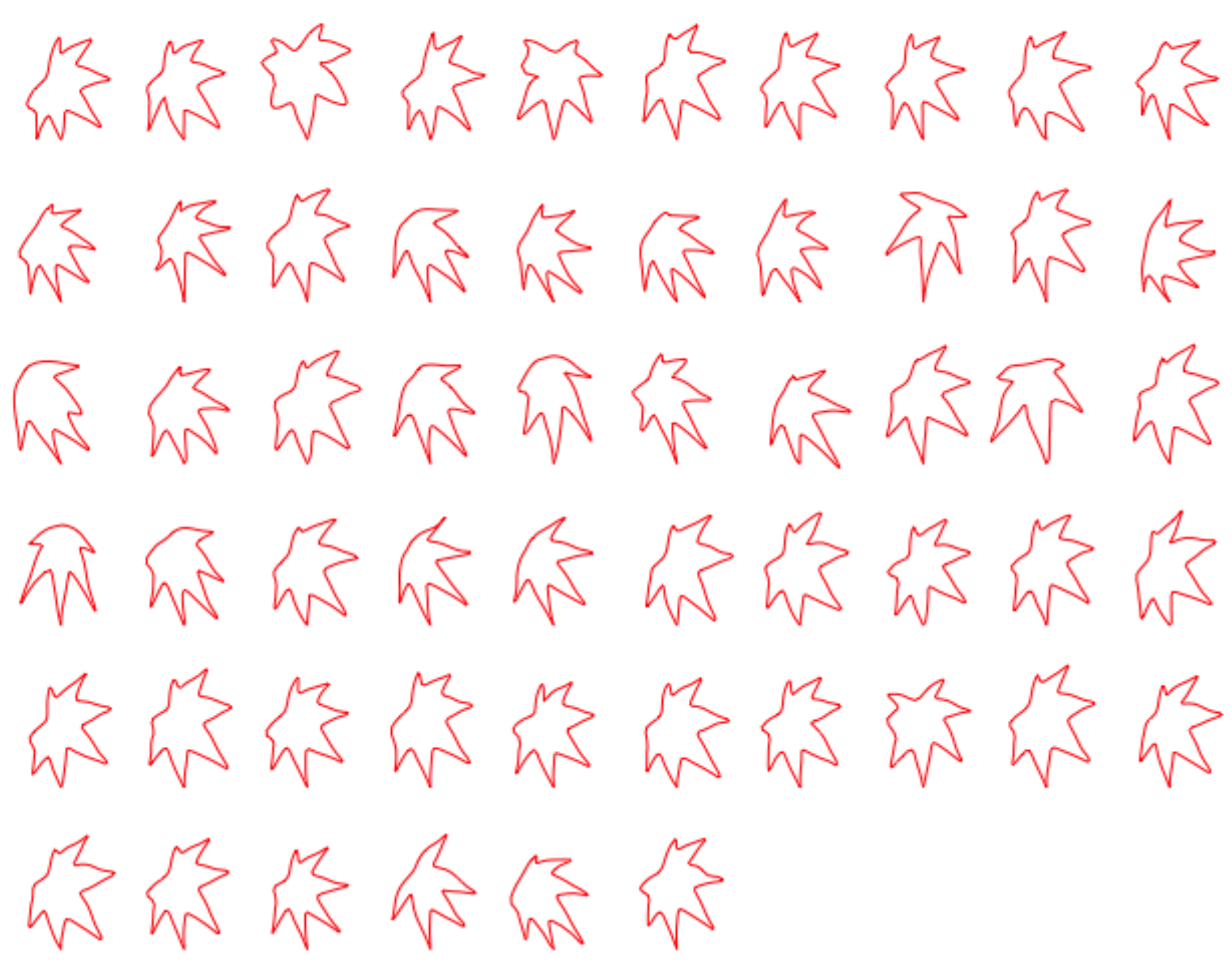}&\includegraphics[scale=.187]{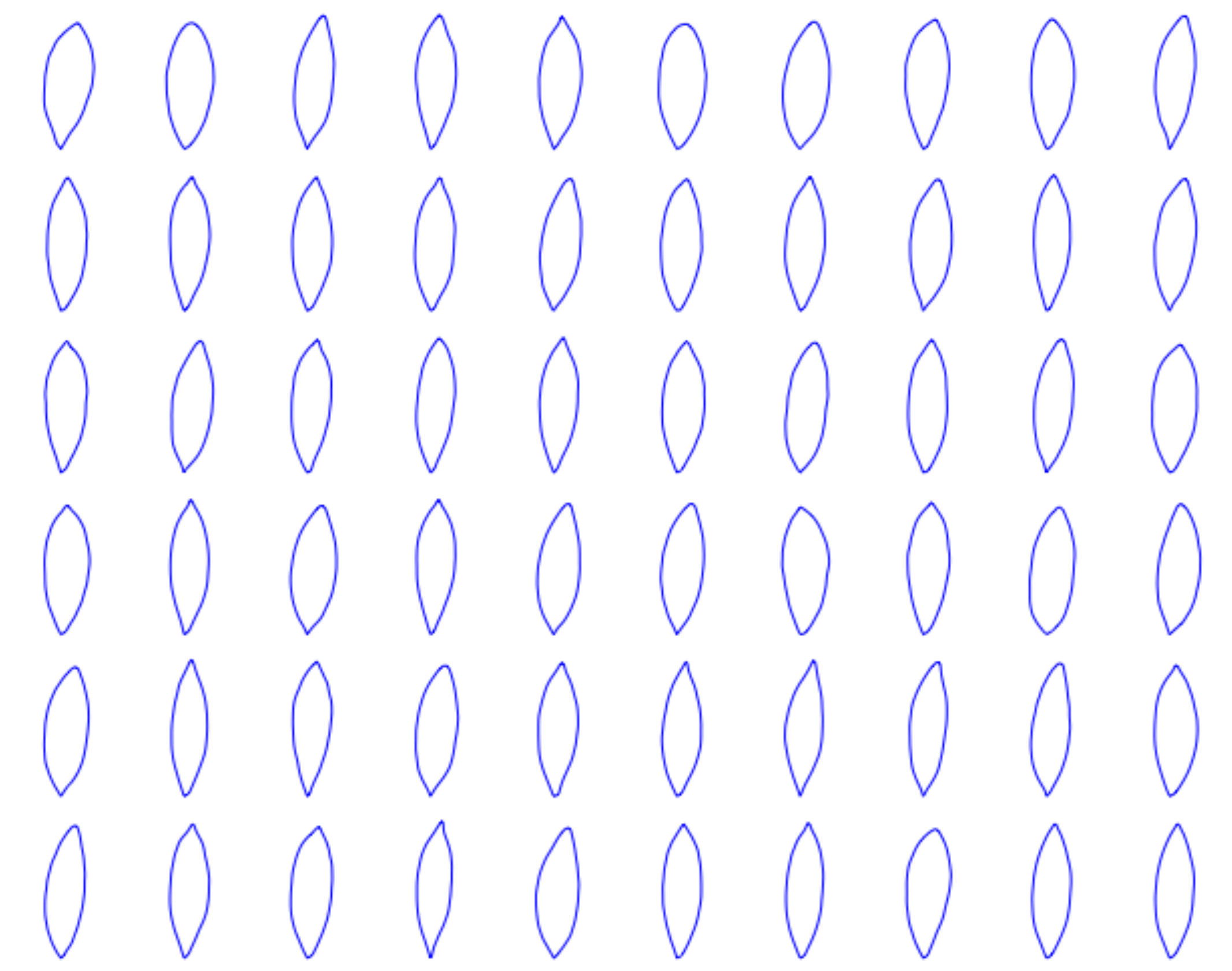}&\includegraphics[scale=.187]{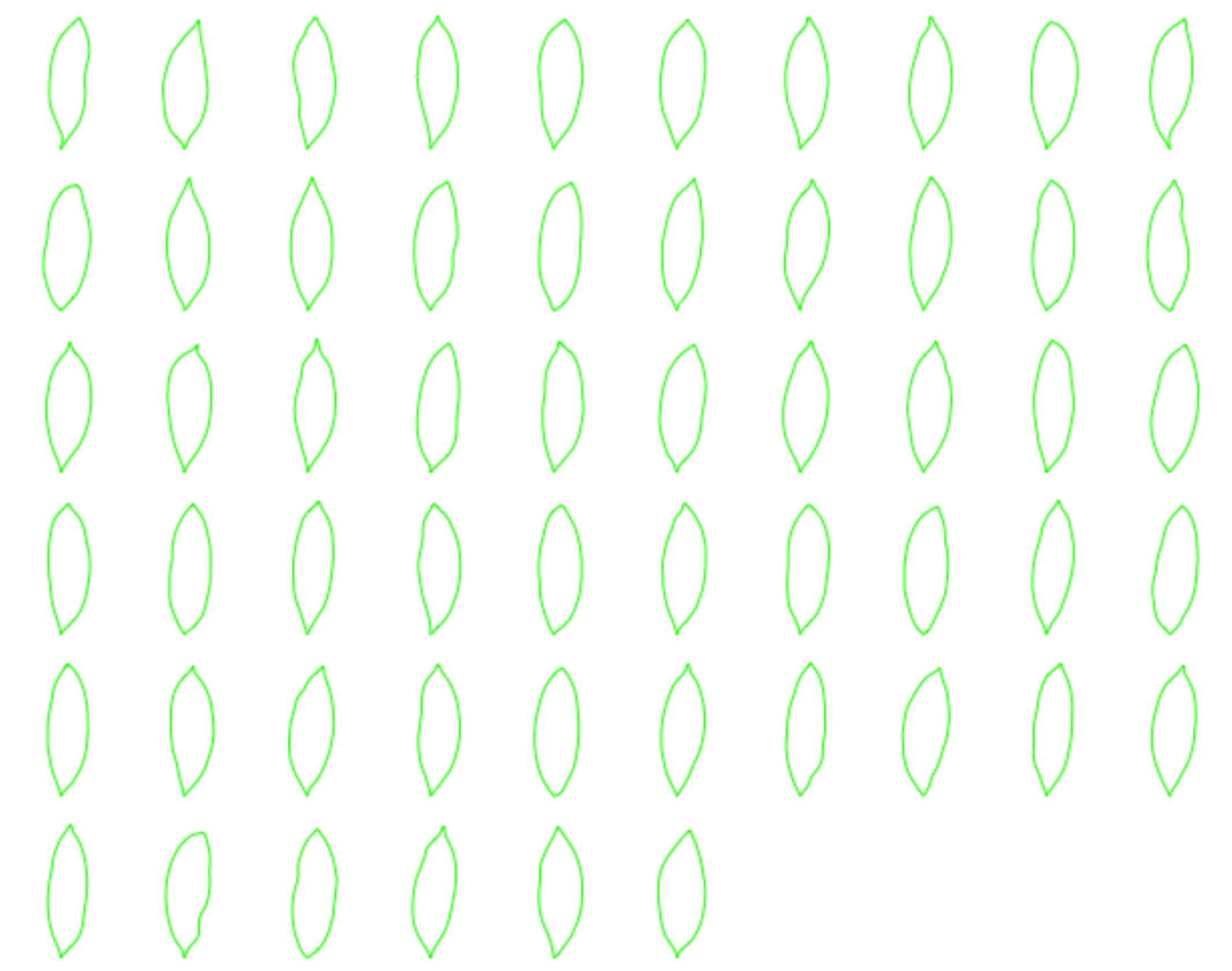}\\
\vspace{-.25in}
\end{tabular}
\includegraphics[scale=1.5]{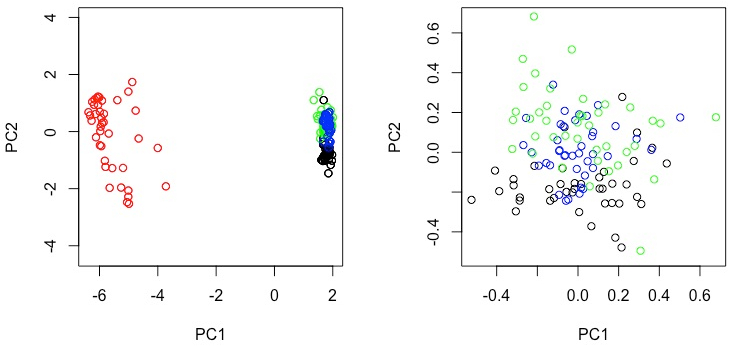}
\end{center}
\vspace{-.25in}
    \caption{Top: Sample leaf shapes from four different plant species. Classes 1 (black), 3 (blue) and 4 (green) are similar. Class 2 (red) is easily discernible. Bottom: Plots of the first two PCs in the tangent space at the mean of all four classes (left) and at the mean of the three similar classes (right). Each point represents one leaf.}
    \label{fig:leaves}
\end{figure}

We consider classification of a small subset of leaves from the Flavia Plant Leaf dataset. To highlight the benefits of the proposed methods, we have selected three classes that are difficult to distinguish (classes 1, 3 and 4 drawn in Figure \ref{fig:leaves} in black, blue and green, respectively) and one outgroup that is easily distinguishable from the other three classes (class 2 drawn in Figure \ref{fig:leaves} in red). To assess classification performance, the dataset is randomly split into a training set of 40 leaves from each class, and a test set consisting of the remaining leaves (23 in class 1, 16 in class 2, 20 in class 3 and 16 in class 4). The first two PCs for one training dataset for all classes are plotted in the bottom-left panel of Figure~\ref{fig:leaves}. It is evident that class 2 (red points) is very far from the other classes and is easily distinguishable using any reasonable classification method, including LDA and QDA. However, classification among the other three classes is much more challenging as there is considerable overlap between the classes (bottom-right panel of Figure~\ref{fig:leaves}).

\begin{table}[!t]
\caption{Average misclassification rate (\%) of LDA and QDA based on 25 different random splits of the data. We consider various points of projection for the three pairwise problems for classes 1, 3 and 4. The pairwise mean shape projection point results are highlighted in bold. Projections involving the outgroup are highlighted in red.}
\vspace{-.15in}
\label{tab:proj}
\begin{center}
\begin{scriptsize}
\begin{tabular}{ccc|ccc|ccc}
\hline
\hline
\multicolumn{3}{c}{Class 1 vs. Class 3} & \multicolumn{3}{|c}{Class 1 vs. Class 4} & \multicolumn{3}{|c}{Class 3 vs. Class 4} \\
\hline
Projection & QDA & LDA & Projection & QDA & LDA & Projection & QDA & LDA \\
\hline
$\bar{q}_{1,4}$ 	& 1.73 & 1.70 & \boldmath$\bar{q}_{1,4}$	& {\bf 5.08} &\bf 4.91 & \boldmath$\bar{q}_{3,4}$& {\bf 4.93} &\bf 4.47 \\
$\bar{q}_{1}$ 	& 2.26 & 2.50 & $\bar{q}_{1,3,4}$& 6.64 & 6.22 & $\bar{q}_{4}$	& 5.30 & 5.05 \\
\boldmath$\bar{q}_{1,3}$ & {\bf 2.75} &\bf 3.12 & $\bar{q}_{3}$	& 6.66 & 15.04 & $\bar{q}_{1,4}$	& 5.47 & 6.63 \\
{\color{red} $\bar{q}_{1,2}$} 	& 2.97 & 3.64 & $\bar{q}_{1}$	& 7.11 & 6.95 & $\bar{q}_{1}$	& 6.53 & 6.90 \\
$\bar{q}_{4}$ 	& 2.99 & 10.51 & $\bar{q}_{1,3}$	& 7.18 & 7.72 & $\bar{q}_{3}$	& 6.88 & 6.75 \\
$\bar{q}_{3}$ 	& 3.31 & 21.44 & $\bar{q}_{3,4}$	& 7.23 & 11.70 & $\bar{q}_{1,3,4}$& 8.26 & 8.59 \\
$\bar{q}_{3,4}$ 	& 3.73 & 15.62 & $\bar{q}_{4}$ 	& 7.64 & 13.75 & $\bar{q}_{1,3}$	& 8.83 & 10.05 \\
$\bar{q}_{1,3,4}$& 4.14 & 4.01 & {\color{red} $\bar{q}$}	& 7.93 & 8.15 & {\color{red} $\bar{q}$}		& 10.00 & 11.33 \\
{\color{red} $\bar{q}_{2,4}$} 	& 4.54 & 15.31 & {\color{red} $\bar{q}_{1,2}$}	& 8.35 & 8.68 & {\color{red} $\bar{q}_{2,3}$}	& 12.09 & 11.22 \\
{\color{red} $\bar{q}$} 		& 4.95 & 6.06 & {\color{red} $\bar{q}_{2,4}$}	& 10.14 & 22.46 & {\color{red} $\bar{q}_{2,4}$}	& 12.41 & 12.70 \\
{\color{red} $\bar{q}_{2,3}$} 	& 7.05 & 17.57 & {\color{red} $\bar{q}_{2,3}$}	& 10.71 & 17.12 & {\color{red} $\bar{q}_{1,2}$}	& 13.11 & 16.52 \\
{\color{red} $\bar{q}_{2}$}  	& 12.42 & 24.93 & {\color{red} $\bar{q}_{2}$}	& 18.27 & 26.66 & {\color{red} $\bar{q}_{2}$}	& 20.16 & 19.83 \\
\hline
\end{tabular}
\end{scriptsize}
\end{center}
\end{table}

Table~\ref{tab:proj} illustrates the impact of the projection point on the misclassification rate. We provide results for pairwise classification for classes 1, 3 and 4 only, and note that classification involving class 2 is always very good. The table reports average misclassification rates, computed over 25 random splits of the data, for 12 different points of projection. We additionally average the misclassification rates over the different numbers of PCs (two through ten) used to define the lower-dimensional space to provide a single performance summary. The candidate projection points are $\bar{q}$ (mean shape of all training samples across the four classes), $\bar{q}_1,\ \bar{q}_2,\ \bar{q}_3,\ \bar{q}_4$ (mean shape of each individual class), $\bar{q}_{1,2},\ \bar{q}_{1,3},\ \bar{q}_{1,4},\ \bar{q}_{2,3},\ \bar{q}_{2,4},\ \bar{q}_{3,4}$ (all pairwise mean shapes), and $\bar{q}_{1,3,4}$ (mean shape across classes 1, 3 and 4). The pairwise mean shape projection point results are highlighted in bold, while projection points involving class 2 are highlighted in red. The results indicate that the pairwise mean shape projection points lead to better classification performance. Furthermore, projection points which include the outgroup in the computation of the mean shape perform poorly. The magnitude of the impact of the projection point on the misclassification rate is striking and is consistent across the three pairwise problems.


\begin{table}[!t]
\caption{Average misclassification rates (\%) of LDA and QDA, for the four leaf species datatset, based on 25 random splits of the data into different training and test sets.}
\vspace{-.15in}
\begin{center}
\begin{scriptsize}
\setlength{\tabcolsep}{5pt}
\label{tab:fourleaf}
\begin{tabular}{c|c|cc|cc|c|cc|cc}
\hline
\hline
& \multicolumn{5}{c}{LDA} & \multicolumn{5}{|c}{QDA} \\  \cline{2-11}
& \multicolumn{3}{|c}{Overall Projection} & \multicolumn{2}{|c}{Pairwise Projections} & \multicolumn{3}{|c}{Overall Projection} & \multicolumn{2}{|c}{Pairwise Projections}\\ \cline{2-11}
& Overall PC & \multicolumn{2}{|c}{Pairwise PCs} & \multicolumn{2}{|c|}{Pairwise PCs} & Overall PC & \multicolumn{2}{|c}{Pairwise PCs} & \multicolumn{2}{|c}{Pairwise PCs} \\ \cline{2-11}
PCs & One Shot & Stage 1  & Stage 3 & Stage 1  & Stage 3 & One Shot & Stage 1  & Stage 3 & Stage 1  & Stage 3\\
\hline
2 &25.97 &18.51 &18.88 &10.99 &11.15 &24.16 &19.31 &18.45 &9.92 &10.77 \\
4 &22.03 &13.33 &11.73 &8.85 &6.19 &18.88 &12.69 &10.29 &6.24 &4.75 \\
6 &17.97 &9.92 &9.49 &6.72 &4.80 &14.67 &9.01 &8.05 &5.28 &4.00 \\
8 &16.21 &8.85 &8.16 &5.65 &4.27 &11.25 &7.31 &5.92 &4.43 &4.00 \\
10 &13.33 &8.11 &7.31 &5.23 &4.21 &10.51 &7.04 &5.71 &3.89 &3.09 \\
12 &11.15 &7.25 &6.88 &4.85 &4.48 &8.69 &6.93 &5.07 &3.68 &3.63 \\
14 &9.17 &7.31 &6.56 &4.69 &4.43 &9.49 &7.57 &5.55 &4.53 &3.57 \\
16 &8.00 &7.52 &6.40 &4.53 &4.21 &9.76 &7.57 &6.13 &4.32 &3.84 \\
18 &6.99 &6.67 &6.13 &4.43 &4.37 &10.08 &8.48 &6.88 &4.64 &4.75 \\
20 &6.67 &6.72 &6.51 &4.43 &4.48 &10.24 &9.23 &7.84 &5.44 &6.03 \\
\hline
\end{tabular}
\end{scriptsize}
\end{center}
\end{table}

Next, we apply the various procedures described in Section~\ref{sec:classify} to 25 different random splits of the data. We consider both LDA and QDA in PC spaces of different dimension ranging from two to 20. Table~\ref{tab:fourleaf} presents the average misclassification rates on the test data. As a general trend, the misclassification rate when using LDA decreases as the number of PCs increases. When using QDA, the misclassification rate decreases as the number of PCs increases from two to ten and then increases slightly. These patterns show the interplay between dimension reduction and the complexity of the model being fit. Overall, QDA performs better than LDA, although LDA performs well for the PP approach with an OS decision for classification (Pairwise Projections, Pairwise PCs and Stage 1). The standard approach SS-OS (Overall Projection, Overall PC, One Shot) provides the poorest performance among the different methods. 

Table~\ref{tab:fourleaf} allows us to assess the value of various components of the procedures we have described: choice of projection point (single or pairwise), choice of PC space (single or pairwise), and choice of decision for classification (one shot or recursive). The table shows that the use of pairwise projection points is beneficial in classification based on both LDA and QDA. The comparison of the One Shot columns to the Stage 1 columns shows the value of selecting pairwise PC spaces rather than performing classification in a single PC space based on all training data. Finally, the value of recursion over the one shot approach is demonstrated by the reduction in misclassification rate from Stage 1 to Stage 3.


\subsection{Classification of the Entire Leaf Dataset}

We now apply all of the proposed procedures to the entire leaf dataset. Since the 32 species of leaves have various shapes, i.e., some classes are very similar while others are very different, we are again interested in investigating the differences in misclassification rate across the various modeling and final decision choices provided by the proposed methods. 
As in Section \ref{sec:fourleaves}, we use 25 different random splits of the data into training and test sets. We use 40 training samples from each class, and the remaining samples for testing. This results in a balanced training set but an unbalanced test set. The total number of test cases across all classes is 627.

\begin{table}[!t]
\caption{Average misclassification rate (\%) of LDA and QDA, for the entire leaf dataset of 32 species, based on a 25 random splits of the data into training and test sets.}
\vspace{-.15in}
\label{tab:allleaf}
\begin{center}
\begin{scriptsize}
\setlength{\tabcolsep}{5pt}
\begin{tabular}{c|c|cc|cc|c|cc|cc}
\hline
\hline
& \multicolumn{5}{c}{LDA} & \multicolumn{5}{|c}{QDA} \\  \cline{2-11}
& \multicolumn{3}{|c}{Overall Projection} & \multicolumn{2}{|c}{Pairwise Projections} & \multicolumn{3}{|c}{Overall Projection} & \multicolumn{2}{|c}{Pairwise Projections}\\ \cline{2-11}
& Overall PC & \multicolumn{2}{|c}{Pairwise PCs} & \multicolumn{2}{|c|}{Pairwise PCs} & Overall PC & \multicolumn{2}{|c}{Pairwise PCs} & \multicolumn{2}{|c}{Pairwise PCs} \\ \cline{2-11}
PCs & One shot & Stage 1  & Stage 31 & Stage 1  & Stage 31 & One shot & Stage 1  & Stage 31 & Stage 1  & Stage 31\\
\hline
2 &54.50 &33.60 &33.72 &24.04 &23.30 &45.69 &24.73 &30.33 &18.09 &19.50 \\
4 &41.72 &31.20 &27.60 &21.79 &19.42 &33.35 &18.01 &22.02 &13.45 &14.36 \\
6 &32.54 &29.14 &25.93 &20.26 &17.91 &24.38 &15.87 &19.27 &10.51 &12.57 \\
8 &30.39 &27.73 &25.06 &18.62 &17.23 &22.41 &14.87 &17.42 &9.73 & 11.88 \\
10 &29.22 &26.67 &24.58 &17.71 &16.96 &21.38 &15.05 &17.58 &9.95 &12.01 \\
12 &27.30 &26.02 &24.52 &17.23 &16.87 &20.34 &15.62 &17.86 &10.53 &12.53 \\
14 &26.18 &25.42 &24.54 &16.89 &16.82 &20.85 &16.37 &18.71 &11.32 &13.29 \\
16 &25.37 &24.91 &24.43 &16.56 &16.82 &22.21 &17.70 &19.37 &12.27 &14.35 \\
18 &24.78 &24.71 &24.41 &16.33 &16.81 &23.77 &19.18 &20.21 &13.22 &15.41 \\
20 &24.57 &24.38 &24.36 &16.08 &16.82 &26.08 &20.89 &21.75 &14.53 &16.81 \\
\hline
\end{tabular}
\end{scriptsize}
\end{center}
\end{table}

Table~\ref{tab:allleaf} shows the average misclassification rates for LDA and QDA. In general, the misclassification rates are larger than those in Table~\ref{tab:fourleaf}, as expected. However, the proposed methods perform quite well, with PP-OS in an eight-dimensional PC space providing the lowest misclassification rate of only $9.73\%$ based on the QDA model. We note that methods that use pairwise projection points and pairwise PC spaces perform better than their single projection counterparts. The LDA misclassification rates decrease as the dimensionality of the PC spaces increases. For QDA, the misclassification rates decrease up to a point and then increase slightly. These trends are the same as those observed in the previous section. Overall, QDA performs better than LDA across all methods. We note that the full recursion does not always perform better than the one shot method in this case. For LDA, it reduces the misclassification rate up to around a 14-dimensional PC space. For QDA, the one shot method always performs better. A careful examination of intermediate stages (not included in Table \ref{tab:allleaf}) suggests that the recursion may help up to a stage where all outgroups are removed from the data. After that, the log-likelihoods exhibit greater numerical stability and averaging across linearizations provides a modest benefit.

\subsection{Classification of the Animal Dataset}

This last set of results considers a dataset of various animals, which poses some additional challenges for shape classification. While the shapes of leaves within the same class were very similar, within class variability for the animal shapes is much larger due to very different poses in which the animals were imaged. We use 25 different random splits of the data into training and test sets of sizes 60 and 40, respectively. Thus, the total number of test cases across all classes is 800.

\begin{table}[!t]
\caption{Average misclassification rate (\%) of LDA and QDA, for the animal dataset, based on 25 random splits of the data into training and test sets.}
\vspace{-.15in}
\label{tab:animal}
\begin{center}
\begin{scriptsize}
\setlength{\tabcolsep}{5pt}
\begin{tabular}{c|c|cc|cc|c|cc|cc}
\hline
\hline
& \multicolumn{5}{c}{LDA} & \multicolumn{5}{|c}{QDA} \\  \cline{2-11}
& \multicolumn{3}{|c}{Overall Projection} & \multicolumn{2}{|c}{Pairwise Projections} & \multicolumn{3}{|c}{Overall Projection} & \multicolumn{2}{|c}{Pairwise Projections}\\ \cline{2-11}
& Overall PC & \multicolumn{2}{|c}{Pairwise PCs} & \multicolumn{2}{|c|}{Pairwise PCs} & Overall PC & \multicolumn{2}{|c}{Pairwise PCs} & \multicolumn{2}{|c}{Pairwise PCs} \\ \cline{2-11}
PCs & One shot & Stage 1  & Stage 19 & Stage 1  & Stage 19 & One shot & Stage 1  & Stage 19 & Stage 1  & Stage 19\\
\hline
12 &62.99 &61.52 &59.54 &45.67 &53.49 &54.58 &48.77 &49.05 &34.40 &41.60 \\
14 &61.17 &60.59 &58.90 &44.50 &52.75 &52.95 &47.68 &48.07 &33.28 &41.14 \\
16 &60.51 &59.85 &58.20 &43.41 &51.87 &51.67 &47.10 &47.45 &32.68 &40.48 \\
18 &59.85 &59.22 &57.47 &42.24 &51.14 &50.64 &46.53 &46.38 &32.39 &40.21 \\
20 &59.08 &58.40 &56.58 &41.37 &50.08 &50.49 &46.21 &46.13 &32.31 &40.49 \\
22 &58.17 &57.69 &56.29 &40.55 &49.35 &50.58 &46.06 &46.27 &32.40 &40.84 \\
24 &58.14 &57.07 &55.95 &39.82 &48.97 &50.63 &46.47 &46.41 &32.84 &41.58 \\
26 &57.49 &56.54 &55.69 &39.16 &48.63 &50.85 &46.87 &46.82 &33.53 &42.32 \\
28 &56.96 &56.15 &55.43 &38.61 &48.37 &51.40 &47.34 &47.38 &34.48 &43.74 \\
30 &56.47 &55.83 &55.28 &37.95 &47.98 &51.85 &48.12 &47.88 &35.57 &44.82 \\
\hline
\end{tabular}
\end{scriptsize}
\end{center}
\end{table}

Table~\ref{tab:animal} shows the average misclassifiction rates for this example. The overall patterns are very similar to the leaf dataset example. However, the misclassification rates in this case are much worse. Overall, QDA performs better than LDA, and the PP procedure leads to smaller misclassification rates than the SP method. The baseline SS method is the worst, as before. The recursion helps only for the SP method and hurts for the PP methods. One thing of note is that many more PCs are required in this example to achieve good classification performance, a result of larger variability within and across classes.


\section{Discussion}
\label{sec:discuss}

An important step in shape analysis is the move from the infinite dimensional, curved space where shapes naturally abide to a finite dimensional linear space that allows the use of a suite of standard statistical tools. In this article, we have shown that this linearization is not trivial, and that the details of the linearization can have a major impact on subsequent statistical inference.

The linearization consists of two main components: a projection point to determine a tangent space and dimension reduction by choice of PCs. We propose aggregation as a mechanism to make use of multiple linearizations driven by different PCs. Aggregation allows us to focus on the pairwise classification problem where the existing literature provides a sound heuristic for the linearization. By itself, the use of pairwise PCs followed by aggregation provides a substantial benefit.

Additionally, we note that the presence of an outgroup can harm the aggregation by contributing linearizations that have little relevance for the classes to which an observation might plausibly be assigned. We propose a recursion that can be quickly computed from the results of the aggregation. The recursion has proven successful for a problem with a modest number of classes and a clear outgroup. For problems with a profusion of classes and great within-class variation, the recursion harms performance.

This work opens up a number of problems, several of which we are in the process of examining. One is the choice of projection points for pairwise comparisons. A sensible alternative to the pairwise mean is the midpoint of the geodesic between the two classwise mean shapes. Although not exactly the same, these two types of projection points are typically close to one another. This will allow us to greatly reduce computational cost: we only have to compute $K$ mean shapes instead of ${K \choose 2}$. A second is whether there are more effective ways to select the PCs for a given projection and pairwise classification problem. A third is to delineate the circumstances under which the recursion is beneficial. A fourth is whether the recursion can be modified to retain aggregation over a subset of linearizations. In future work, we expect to consider these and other problems.\\

\if0\blind
{
\noindent\textbf{Acknowledgements:} We thank Dr. Hamid Laga from Murdoch University for providing the outlines from the Flavia Plant Leaf dataset. This research was partially supported by NSF DMS 1613110 (to SM), and NSF DMS 1613054, NSF CCF 1740761 and NIH R37 CA214955 (to SK).
} \fi

\if1\blind
{
} \fi









\bibliographystyle{chicago}
\bibliography{Bibliography-MH}

\end{document}